\documentclass[unnumsec,webpdf,contemporary,large]{oup-authoring-template}%

\usepackage{textcomp}
\graphicspath{{Fig/}}
\usepackage{subfig}

\theoremstyle{thmstyleone}%
\theoremstyle{thmstyletwo}%
\theoremstyle{thmstylethree}%

\begin{document}

\journaltitle{Briefings in Bioinformatics}
\DOI{DOI HERE}
\copyrightyear{2022}
\pubyear{2022}
\access{Advance Access Publication Date: Day Month Year}
\appnotes{Paper}

\firstpage{1}


\title[Pathology image-based differential gene expression]{Contrastive learning-based computational histopathology predict differential expression of cancer driver genes}

\author[1]{Haojie Huang}
\author[1]{Gongming Zhou}
\author[2]{Xuejun Liu}
\author[1]{Lei Deng}
\author[3]{Chen Wu}
\author[3,$\ast$]{Dachuan Zhang}
\author[2,$\ast$]{Hui Liu}

\authormark{Huang et al.}

\address[1]{\orgdiv{School of Computer Science and Engineering}, \orgname{Central South University},\orgaddress{\postcode{410075}, \state{Changsha}, \country{China}}}
\address[2]{\orgdiv{School of Computer Science and Technology}, \orgname{Nanjing Tech University}, \orgaddress{ \postcode{211816}, \state{Nanjing}, \country{China}}}
\address[3]{\orgdiv{The third affiliated hospital of Soochow University}, \orgaddress{\postcode{213100}, \state{Changzhou}, \country{China}}}

\corresp[$\ast$]{Correspondence should be mainly addressed to Hui Liu: \href{email:hliu@njtech.edu.cn}{hliu@njtech.edu.cn}. Correspondence may also be addressed to \href{email:zhangdachuan@suda.edu.cn}{zhangdachuan@suda.edu.cn.}}

\received{Date}{0}{Year}
\revised{Date}{0}{Year}
\accepted{Date}{0}{Year}



\abstract{\textbf{Motivation:} Digital pathological analysis is run as the main examination used for cancer diagnosis. Recently, deep learning-driven feature extraction from pathology images is able to detect genetic variations and tumor environment, but few studies focus on differential gene expression in tumor cells. \\
\textbf{Results:}  In this paper, we propose a self-supervised contrastive learning framework, HistCode, to infer differential gene expression from whole slide images (WSIs). We leveraged contrastive learning on large-scale unannotated WSIs to derive slide-level histopathological features in latent space, and then transfer it to tumor diagnosis and prediction of differentially expressed cancer driver genes. Our extensive experiments showed that our method outperformed other state-of-the-art models in tumor diagnosis tasks, and also effectively predicted differential gene expression. Interestingly, we found the genes with higher fold change can be more precisely predicted. To intuitively illustrate the ability to extract informative features from pathological images, we spatially visualized the WSIs colored by the attention scores of image tiles. We found that the tumor and necrosis areas were highly consistent with the annotations of experienced pathologists. Moreover, the spatial heatmap generated by lymphocyte-specific gene expression patterns was also consistent with the manually labeled WSIs.
}

 \boxedtext{
 \begin{itemize}
 \item Self-supervised contrastive learning was applied to large-scale unlabeled digital pathology images and extracted tile-level features, which were then aggregated to build the slide-level features via attention pooling. Adversarial negative samples were generated to pose challenge that drove the self-supervised learning to capture informative representation from large-scale unlabeled tiles.
 \item The computational pathological features have been shown highly predictive of both tumor diagnosis and differential gene expressions. Interesting, the prediction accuracy was positively correlated to fold-change level, which indicated that dramatic variation of underlying molecule expression pattern would be more reflected in phenotypic features.
 \item We explored the model interpretability via spatial deconvolution, and colored each tile according to its normalized attention scores. The spatial localization of high attention-scored tiles showed high consistence to the distribution of tumor tissues and immune infiltrating cells annotated by an experienced pathologist.
 \end{itemize}}

\maketitle

\section{Introduction}
With the advancement of scanner and imaging technology, pathology glass slides are increasingly digitized and computational histopathology analysis has emerged as a new standard of diagnostic workflow. Digital histopathology promotes the efficiency and accuracy of pathologists in disease diagnosis and clinical grading.

By virtue of the power of automatic feature extraction, deep learning is increasingly used to extract features from pathology images for multiple subsequent applications. Some deep learning-based models used pathological images to predict the prognosis of cancer patients, such as colorectal cancer\citep{kather2019predicting, bychkovdeep}, hepatocellular carcinoma\citep{saillard2020predicting}, pancreatic ductal adenocarcinoma\citep{ju2021robust} and even panCancer\citep{cheerla2019deep}. Some studies focused on classification of tumor subtypes and treatment efficacy by deep computational pathology analysis\citep{hou2016patch, noorbakhsh2019pan, mobadersany2018predicting, strom2020artificial}. For example, Li et al. proposed a classification model without manual annotations by fusing multi-scale features extract from unlabeled patches tiled from whole slide images\citep{li2021dual}. Saltz et al. studied a variety of cancer types and proposed a computational staining method based on deep learning, which can deconvolve the spatial structure of tumor infiltrating lymphocytes\citep{saltz2018spatial}. Gheisari et al. combined convolution deep belief network and feature encoding from WSIs to classify neuroblastoma\citep{gheisari2018convolutional}.

The aforementioned studies promoted the application of digital pathology to routine clinical diagnosis and prognosis. However, these studies were limited to the diagnostic and prognostic value, the potential of digital pathological images has not been fully mined. The development of high-throughput sequencing technology generated large-scale multi-omics datasets. Based on the assumption that specific micromolecular patterns will be reflected in histological morphology and cell phenotype, some studies begun the exploration of pathological image features to infer molecular patterns. For example, Kather et al. used deep neural network to predict a wide range of gene mutations from histology images, which verified and quantified the relationship between genotype and phenotype in cancer cells\citep{kather2020pan}. Yu et al. found that, in most cancers, there is a general association between histological morphology features and genetic mutations\citep{fu2020pan}. Chen et al. combined the histopathology features and molecular patterns to improve the diagnosis and prognosis of cancer patients\citep{chen2020pathomic}. Coudray et al. further predicted gene mutations based on the histopathological features derived for the classification task of non-small cell lung cancer subtypes by using deep convolution network\citep{coudray2018classification}. In addition, a lot of studies have demonstrated the applicability of computational pathology to predict microsatellite instability\citep{kather2019deep, echle2020clinical, cao2020development, yamashita2021deep} and detect mitosis\citep{albarqouni2016aggnet, tellez2018whole, rao2018mitos, li2019weakly}.

To further explore the relationship between gene expression profiles and histopathological features, we propose a new framework, HistCode, to exploit the histopathology images by self-supervised contrastive learning to infer the differential gene expression in tumor cells. More precisely, we applied adversarial contrastive learning to extract tile-level features, and then aggregated the features to produce the slide-level latent representation by attention pooling. The slide-level representations were transferred to multiple downstream tasks, including tumor diagnosis and differential gene expression prediction. For systematically evaluating our proposed method, we conducted extensive experiments to demonstrate the expressiveness of computational histopathological features, including tumor diagnosis, quantitative analysis of differential gene expression and spatial localization of lymphocytes. The experimental results fully verify that our model is superior to state-of-the-art models in tumor diagnosis task. We visualized the whole slide images by the attention scores of tiles, and found that the tumor and necrosis area are highly consistent with the annotation of pathological experts, which strongly solidified the capacity of HistCode to extract the informative features from pathological images. Next, we verified the ability of HistCode to predict differential gene expression. Interestingly, we observed that the genes with higher fold change can be more precisely predicted. The activation map generated by lymphocyte marker gene expression level was also consistent with the labeled slide by an experienced pathologist. To our best knowledge, we are the first to apply computational pathology for differential gene expression analysis. We believe our work would yield inspiring insights into digital pathological analysis and greatly extend its clinical application.

\section{Materials and methods}
\subsubsection{Whole slide image}
All hematoxylin and eosin-stained digital slides (frozen tissue) were downloaded from TCGA via the Genomic Data Commons Data Portal. We collected two solid tumors, breast and lung cancer, in our study, because there are enough samples for us to establish solid model, as well as for systematic evaluation. From TCGA-BRCA project, we gathered 1,979 WSIs from 1,094 breast cancer patients, including 1,580 tumor samples and 399 normal samples. We also collected the breast cancer slides from CPTAC-BRCA project, which had 642 WSIs from 134 participants. This CPTAC-BRCA dataset was used as an independent test set to verify the generalizability of our model. The lung cancer dataset comes from the project TCGA-LUSC and TCGA-LUAD. There are in total 2,168 WSIs from 1,010 patients, including 1,577 tumor WSIs and 591 normal WSIs.

Only the slide with a magnification greater than 20* were included in this study. The slide-level diagnosis provided by TCGA database were used as ground truth for classification labels.

\subsubsection{Slide annotation}
A pathologist with more than ten-year experience was asked to manually annotate a few slides. The annotated slides were used for the validation of spatial localization of tumor and necrosis area, as well as tumor infiltrating lymphocytes. During annotation, the pathologist was blinded with regard to any molecular or clinical feature.

\subsubsection{RNA-seq dataset}
The RNA-seq datasets from TCGA are used for differential expression analysis. We selected the patients with both tumor and normal pathology images, and obtained 153 matched samples for breast cancer.  Genome-wide expression profiles in fragment per kilobase million with upper-quartile normalization (FPKM-UQ) were used for differential gene expression analysis.

\subsubsection{Cancer driver genes}
In this study, we focused on the cancer driven genes, as their differential expression is the drive force for the generation and proliferation of tumor cells. The driver genes is obtained from the compendium of cancer driver genes curated by Mart{\'\i}nez-Jim{\'e}nez et al\citep{martinez2020compendium}. From the total of 568 driver genes, we chose the ones that has been reported high cumulative number of non-synonymous mutations. As a result, we filtered out top 200 most mutational driver genes included in our differential expression analysis.

\subsection{Image processing}
Due to the high dimensionality of the whole slide images (up to 100,000$\times$100,000 pixels), we tessellated slides into small-size squares (tiles) so that they are ready as input of deep learning model. Before tiling, we read slides using python library openslide\citep{goode2013openslide} and used the Otsu algorithm\citep{otsu1979threshold} to select areas that contained enough tissue cells. The threshold 8 was used to detect tissue. After exclusion of white background, the slide was tessellated to 128$\times$128\textmu m (256$\times$256 pixels) tiles. Of note, we only stacked the coordinates of each tile and the slide metadata using the hdf5 hierarchical data format. Next, those tiles where the surface area covered by tissue cells is less than 100 were removed. Finally, we got 14,705,914 tiles from TCGA-BRCA slides, 5,888,085 tiles from CPTAC-BRCA slides, and 12,769,300 tiles from TCGA-Lung slides. To eliminate the influence of different staining condition and imaging protocol of different WSI datasets, we applied color normalization on tiles.

\subsection{Differential expression analysis}
We calculated the fold change of each gene in tumor relative to normal tissues. Because the absolute expression levels of different genes vary in large scope, especially when the gene expression level of normal tissue is close to 0, fold change give rise to instability. To overcome this problem, we scaled the fold change by $\log_{10}(fc + 1)$\citep{schmauch2020deep} for each gene per patient.
Apart from the quantitative prediction of differential expression level, we also cast the problem into binary classification task. If the $|\log_2(fc)|$ is more than 1.5\citep{liao2019identification}, the label of corresponding gene was set to 1, and 0 otherwise.

\subsubsection{HistCode framework}
\begin{figure*}[htb]
	\centering
	\includegraphics[scale=0.8]{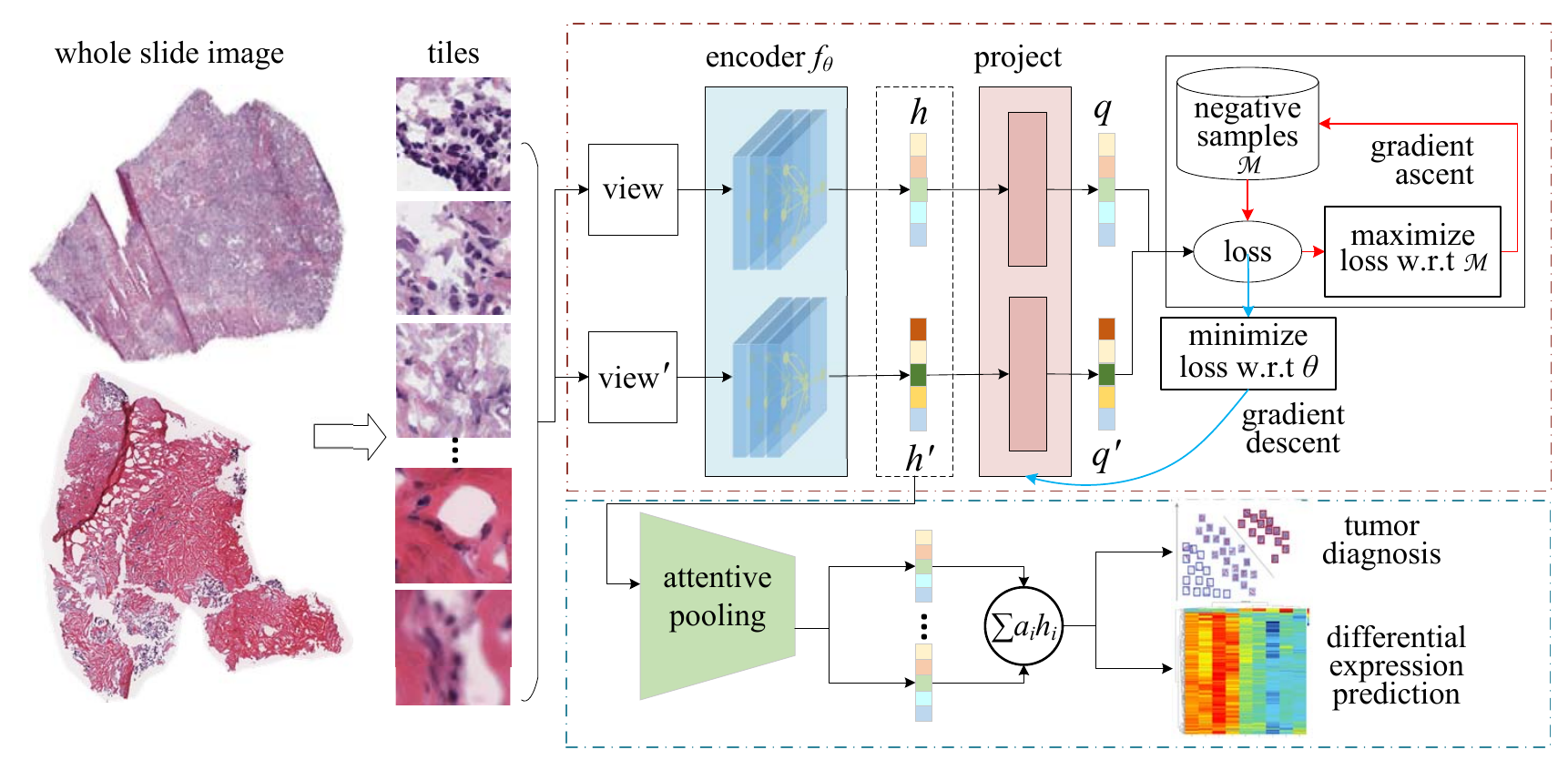}
	\caption{Illustrative flowchart of HistCode framework. There were three steps, slide proprocessing, contrastive learning-based pretraining (dark red box) transfer learning to downstream tasks (blue box). The cyan lines represented backward propagation messages, and red lines represented the generation process of adversarial negative samples.}
	\label{fig:flowchart}
\end{figure*}

Our HistCode framework included three modules, as shown in Figure~\ref{fig:flowchart}. First, the self-supervised contrastive learning is used to learn latent representation of tiles. Next, we leveraged the gated-attention pooling to aggregate the tile-level features to build slide-level features. In the downstream tasks, we transferred the slide-level features to tumor diagnosis and differential gene expression prediction.

\subsubsection{Contrastive learning for feature extraction}
As we have only slide-level labels (tumor or normal), supervised learning is not applicable to extract features of tiles. We adopted self-supervised pretraining on large-scale unannotated tiles to obtain tile-level features. In this study, the adversarial contrastive learning, AdCo\citep{hu2021adco}, is adopted for pretraining. We have also tested another contrastive learning methods, SimCLR, and found AdCo achieved better performance in downstream tasks.

Formally, denote the input tile as by $x_i$, we used CNN as the backbone network $f_{\theta}$ to transform tiles into latent representation $h_i=f_{\theta}(x_i)$, which is then projected into embedding $q_i$ by a multi-layer projection. The contrastive learning is to train the network parameter $\theta$ to discriminate query sample $q_i$ from a set $K$ of negative samples. The contrastive loss was defined as:

\begin{equation}
	L=-\frac{1}{N}\sum_{i=1}^N log\frac{exp(q_i qi^{'}/\tau)}{exp(q_i q_i^{'}/\tau)+\sum_{k=1}^K exp(q_i m_k/\tau)}
	\label{eq:loss}
\end{equation}

where $q'_i$ is the embedding of augmentation of the same instance $x_i$, which is considered as the positive sample for the query $q_i$, and $\tau$ is a positive value of temperature.

Inspired from adversarial learning, AdCo aims to generate challenging negative samples to be distinguished from query samples. Mathematically, AdCo takes the embedding of all samples as negative adversaries and updates the negative samples to maximize the contrastive loss, so that the adversarial negative samples were push closer to query samples. A memory bank is maintained to stack the negative samples. So, we have the following min-max problem:
\begin{equation}
    \theta^{*},\mathcal{M}^{*}=\arg \min_{\theta} \max_{\mathcal{M}} L(\theta, \mathcal{M})
	\label{eq:arg}
\end{equation}
in which $\mathcal{M}$ represents the set of dynamically updated adversarial negative samples, which can be regarded as a set of model parameter $\mathcal{M}$. During the training process, the network parameter $\theta$ is updated along the descending direction of the gradient, while $M$ is updated along the ascending direction of the gradient. So, the parameter $\theta$ and $M$ is alternately updated as below:
\begin{equation}
    \theta \leftarrow \theta-\lambda_{\theta}\frac{\partial L(\theta, \mathcal{M})}{\partial \theta}
	\label{eq:theta}
\end{equation}
\begin{equation}
    m_k \leftarrow m_k-\lambda_{m_k}\frac{\partial L(\theta, \mathcal{M})}{\partial m_k}
	\label{eq:M}
\end{equation}

As the size of total tiles is too large to be loaded into memory, we randomly select a percentage of tiles from each slide and stacked them in the memory bank. This is reasonable because pathology tiles has relative low information density compared to natural picture. The final goal of the solution is to achieve the saddle point. The adversarial negative samples forced the encoder to capture essential information of each image to discriminate it from others.

In our implementation, ResNet50 was used as the backbone network. We used only the first four main layers (the output of the last layer is 1024), and load the pre-trained weights on ImageNet. Two fully-connected layers were added to the model. In the training process, SGD\citep{bottou2010large} was used as optimizer. The learning rate of the backbone network is set to 0.03. The weight decay is set to 0.0001, and the momentum is set to 0.9.

\subsubsection{Feature aggregation}
For downstream tasks, we need to aggregate the tile-level features to derive the slide-level features. In addition to widely adopted max-pooling and mean-pooling, we also introduced gated-attention pooling\citep{ilse2018attention} to aggregate tile-level features.

There is a trainable aggregation strategy based on attention mechanism.  Let $H =\{h_1, . . . , h_L\}$ be $L$ tile-level embedding  of a slide, the gated-attention pooling is actually the instance-level weighted average pooling:
\begin{equation}
    z=\sum_{i=1}^L a_i h_i
	\label{eq:M}
\end{equation}
in which
\begin{equation}
    a_i=\frac{exp(w^T(\tanh(Vh_i)\bigodot sigm(Uh_i)))}{\sum_{j=1}^L exp(w^T(\tanh(Vh_j)\bigodot sigm(Uh_j)))}
	\label{eq:M}
\end{equation}

where $U$ and $V$ are learnable parameters, $\bigodot$ is an element-wise multiplication and $sigm()$ is the sigmoid non-linearity. The gating mechanism introduces a learnable non-linearity that potentially removes the troublesome linearity in $\tanh$.

\subsubsection{Tumor diagnosis}
Given the learned slide-level representation, we applied transfer learning to conduct downstream tasks. For the tumor diagnosis task, we used a fully connection layer and a softmax layer. The slide features were fed into the fully-connected layer, and the softmax layer output the probability indicating that the input slide included tumor tissues or not. We used argmax to obtain the prediction probability, and then used the cross entropy as loss function to calculate the loss between the slide label $y_i$ and the prediction label $\widehat{y_i}$:
\begin{equation}
    L_{TD} = -\sum_{i=1}^{S}[y_i\log \widehat{y_i}+(1-y_i)\log(1-\widehat{y_i})]
	\label{eq:M}
\end{equation}
in which $S$ is the total number of slide included in the tumor diagnosis task. Note that during the downstream task, only the parameters of the gated-attention pooling and downstream linear layers were updated, while the the parameters in the contrastive learning module were frozen.

\subsubsection{Differential expression prediction}
Since the differential gene expression is actually the change of transcriptional level in tumor cells compared to normal cells, we chose a number of tiles that were most possibly located in tumor and normal tissues to predict differential gene expression. The attention weights learned in the tumor diagnosis task were indicative of whether the tiles contains tumor cells or not, we leveraged the learned attention weights to select tiles. More precisely, we sorted the tiles according their attention weights in descending order, and then selected the $l$ highest tiles and $l$ lowest tiles. In our study, $l$ takes 100. Next, the embedding of the $l$ highest and lowest tiles were averaged, respectively. Finally, the two averaged embeddings were concatenated.
Taking the concatenated embedding as input, we used a linear regression model for each gene to predict its differential expression level. The linear model include only a fully connected layer with 1024 input nodes and 1 output node. The MSE (mean squared error) was used as loss function:
\begin{equation}
    L_{DE} = \frac{1}{Q}\sum_{i=1}^Q (y_i-\widehat{y_i})^2
	\label{eq:M}
\end{equation}

where $y_i$ is the RNA-seq derived differential expression level, and $\widehat{y_i}$ is the predicted one, $Q$ is the total number of genes included in the differential expression task. In our practice, we found that single fully connected layer can achieve good performance in the downstream tasks.

\section{Results}
\subsection{Tumor diagnosis}
For tumor diagnosis, we performed a classification task to predict the slide-level labels. For both breast and lung cancer datasets, we adopted 5-fold cross validation to evaluate model performance. For each fold, the slides were split into three subsets at the level of patients, and 80\% slides were used for training, 10\% for validation and 10\% for testing. The model performance was reported by the mean predicted accuracy on testing set of 5-fold cross-validation.

On TCGA-LUNG cohort, our method achieved accuracy 0.963 and ROC-AUC 0.976. Moreover, we compared our method to three other competitive methods, including MIL-RNN\citep{campanella2019clinical}, ABMIL\citep{ilse2018attention}, DSMIL\citep{li2021dual}. As shown in Table 1, our method outperformed these competitive methods by at least 4\% accuracy. The results show that HistCode successfully extract the features from pathological images for tumor diagnosis.

\begin{table}[!ht]
    \centering
	\caption{Performance comparison of HistCode and competitive methods on TCGA-LUNG cohort}
	\label{Tag_01}
    \begin{tabular}{c|c|c}
    \hline
        Method & Accuracy & AUC   \\ \hline
        MIL-RNN[34] & 0.8619 & 0.9107  \\
        ABMIL[32] & 0.9000 & 0.9551  \\
        DSMIL[35] & 0.9268 & 0.9633  \\ \hline
        Max pooling & 0.8930 & 0.8846  \\
        Mean pooling & 0.9114 & 0.9073  \\
        HistCode & 0.9630 & 0.9765  \\ \hline
    \end{tabular}
\end{table}

\begin{figure*}[htbp]  
	\captionsetup{labelformat=simple, position=top}
	\centering
	\subfloat[ROC curve]
	{
		\begin{minipage}[b]{\columnwidth}
			\centering
			\includegraphics[width=7cm]{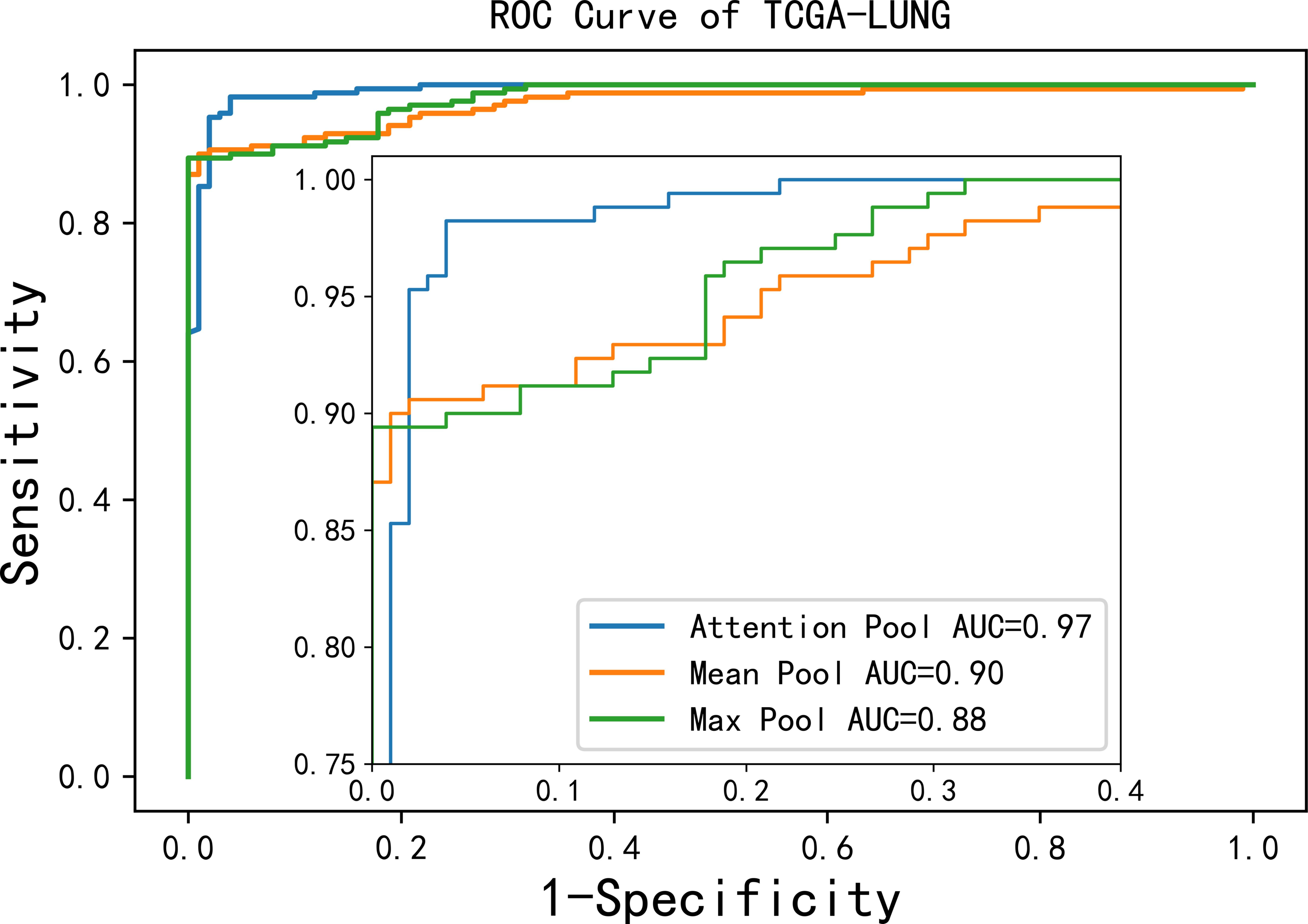}
	\end{minipage}}
	\subfloat[PR curve]
	{
		\begin{minipage}[b]{\columnwidth}
			\centering
			\includegraphics[width=7cm]{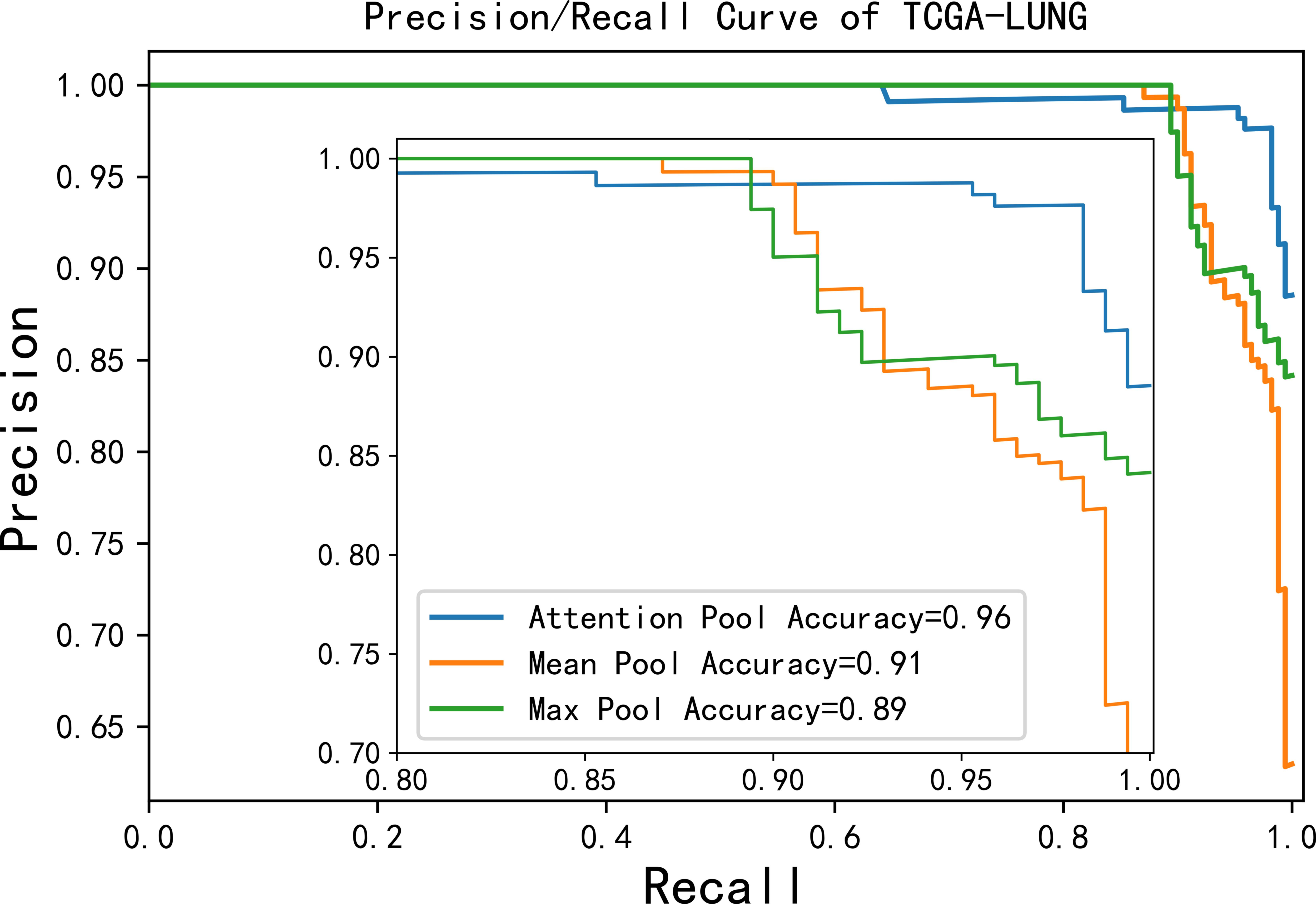}
	\end{minipage}} \\
	\caption{ROC curves and precision-recall curves achieved by our method for tumor diagnosis task on TCGA-LUNG cohort. It shows that gated-attention pooling of tile-level features improves the performance, compared to max-pooling and mean-pooling}
	\label{fig:roc_pr_lung}
\end{figure*}

Also, we explored the impact of different tile-level feature aggregation strategies. we presented the ROC curve and precision-recall curves of three aggregation strategies on the TCGA-LUNG cohort in Figure~\ref{fig:roc_pr_lung}. It can be found that the gated-attention pooling acquired significantly better performance than mean-pooling, while max pooling performed relatively poor.

\begin{figure}[htbp]  
	\captionsetup{labelformat=simple, position=top}
	\centering
	\subfloat[ROC curve]
	{
		\begin{minipage}[b]{.5\columnwidth}
			\centering
			\includegraphics[width=4cm]{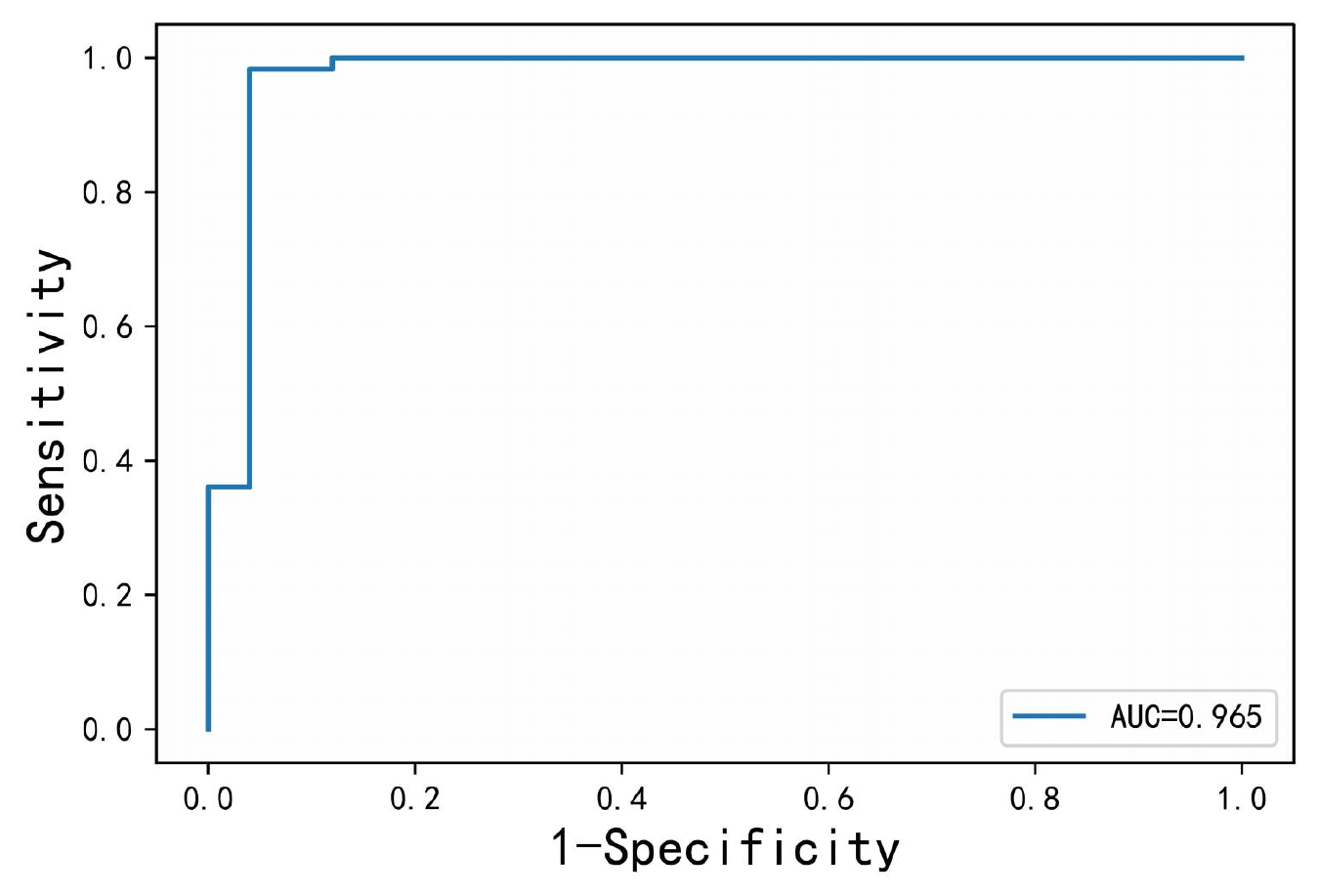}
	\end{minipage}}
	\subfloat[PR curve]
	{
		\begin{minipage}[b]{.5\columnwidth}
			\centering
			\includegraphics[width=4cm]{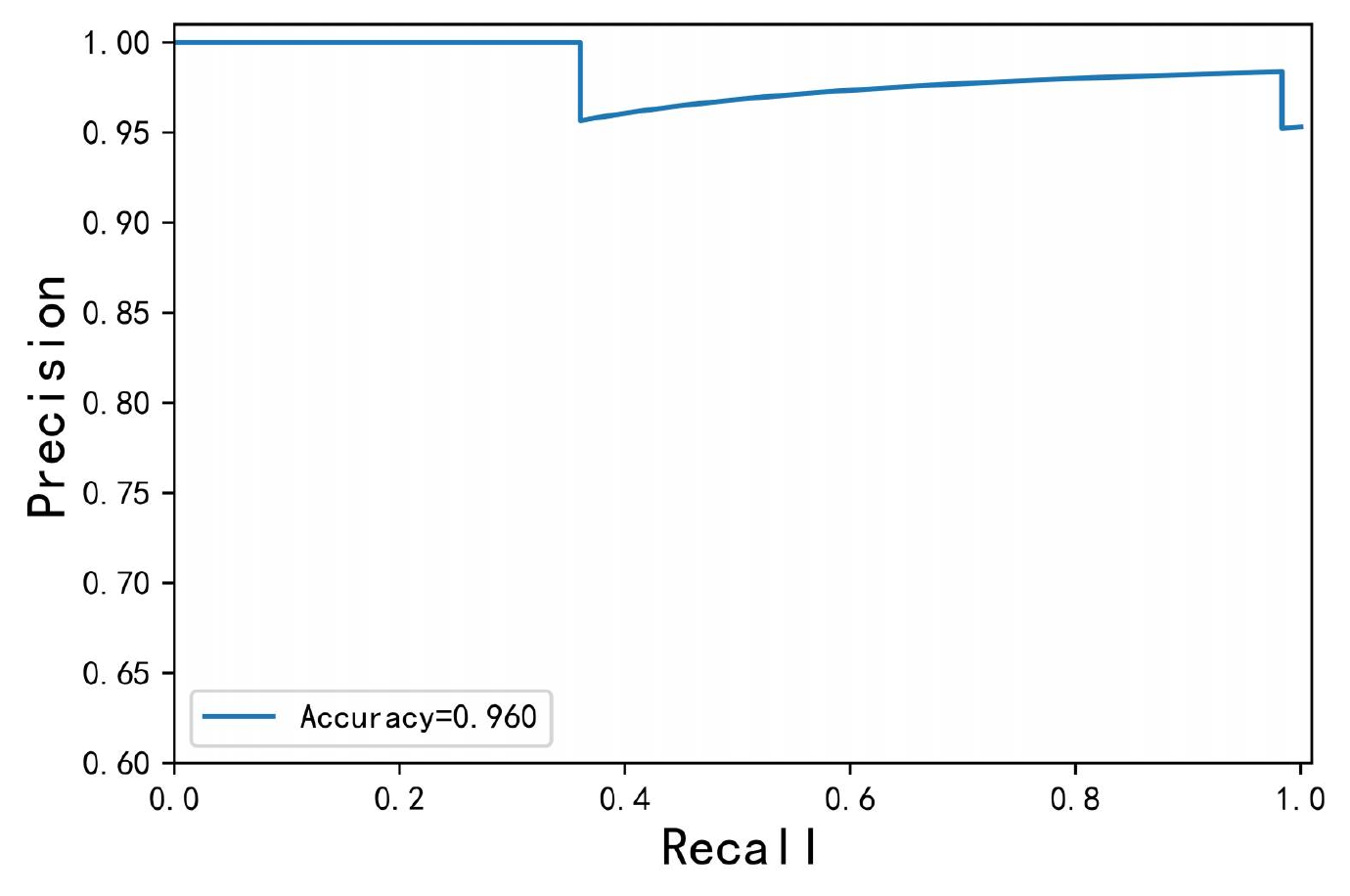}
	\end{minipage}} \\
	\caption{ROC curves and precision-recall curves achieved by our method for tumor diagnosis task on TCGA-BRCA cohort.}
	\label{fig:roc_pr_brca}
\end{figure}

On TCGA-BRCA cohort, our model achieved ROC-AUC 0.965 and accuracy 0.960, as shown in Figure \ref{fig:roc_pr_brca}. To verify the generalizability of our model, we trained the model on TCGA-BRCA cohort, and then used it to predict the slide labels of CPTAC-BRCA cohort. On CPTAC-BRCA cohort, our model obtained ROC-AUC value 0.962 and accuracy 0.931. This result strongly verified the robustness of our method.

\subsection{Differential gene expression prediction}
Based on the attention weights derived from the tumor diagnosis, we chose a number of tiles with highest and lowest attention weights to established the slide-level feature for differential expression prediction (see method), so that the resulting features integrated both tumor and normal tiles.

\begin{figure}[htbp]  
	\captionsetup{labelformat=simple, position=top}
	\centering
	\subfloat[Predicted]
	{
		\begin{minipage}[b]{0.5\columnwidth}
			\centering
			\includegraphics[width=4.2cm]{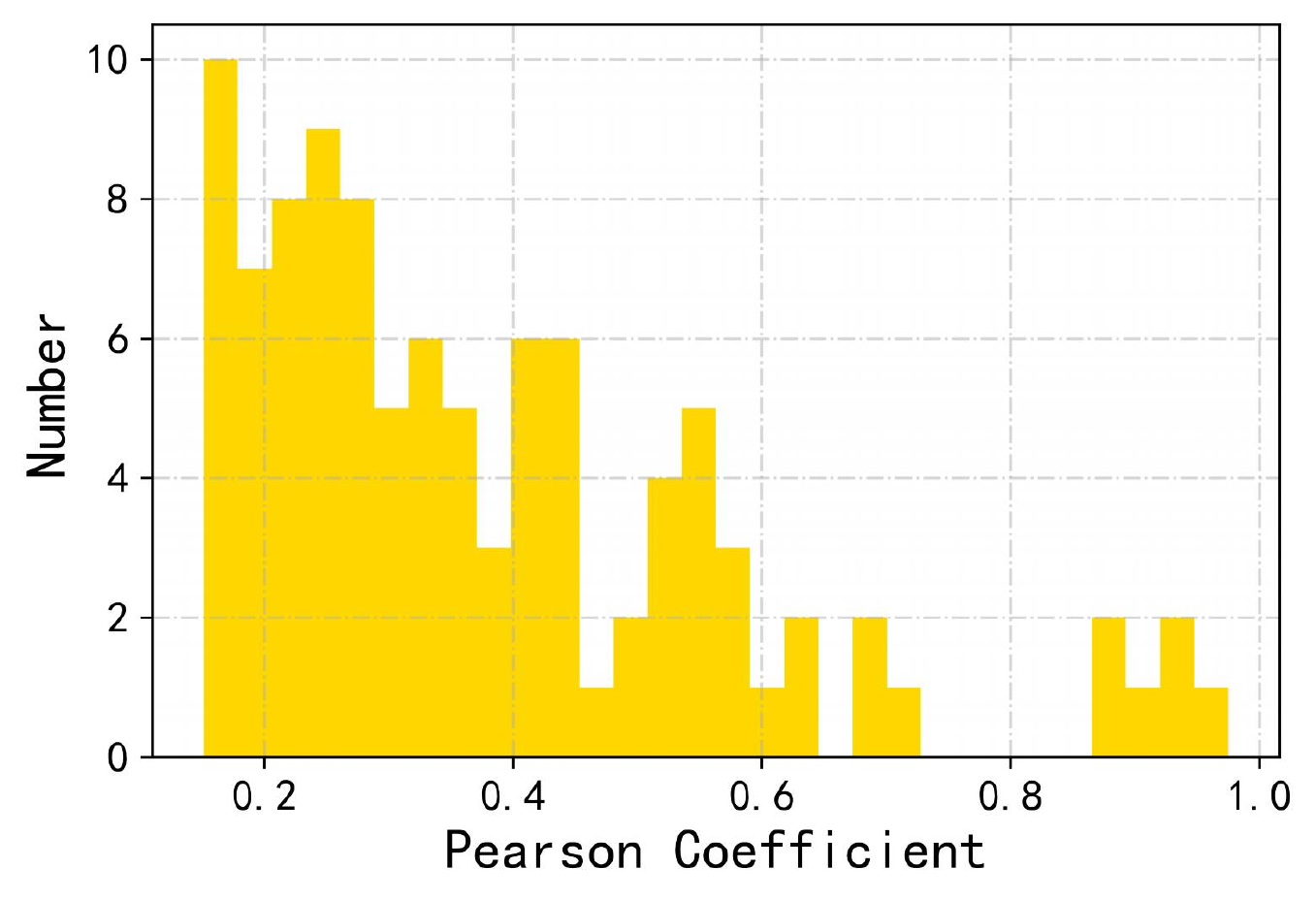}
	\end{minipage}}
	\subfloat[Random]
	{
		\begin{minipage}[b]{0.5\columnwidth}
			\centering
			\includegraphics[width=4cm]{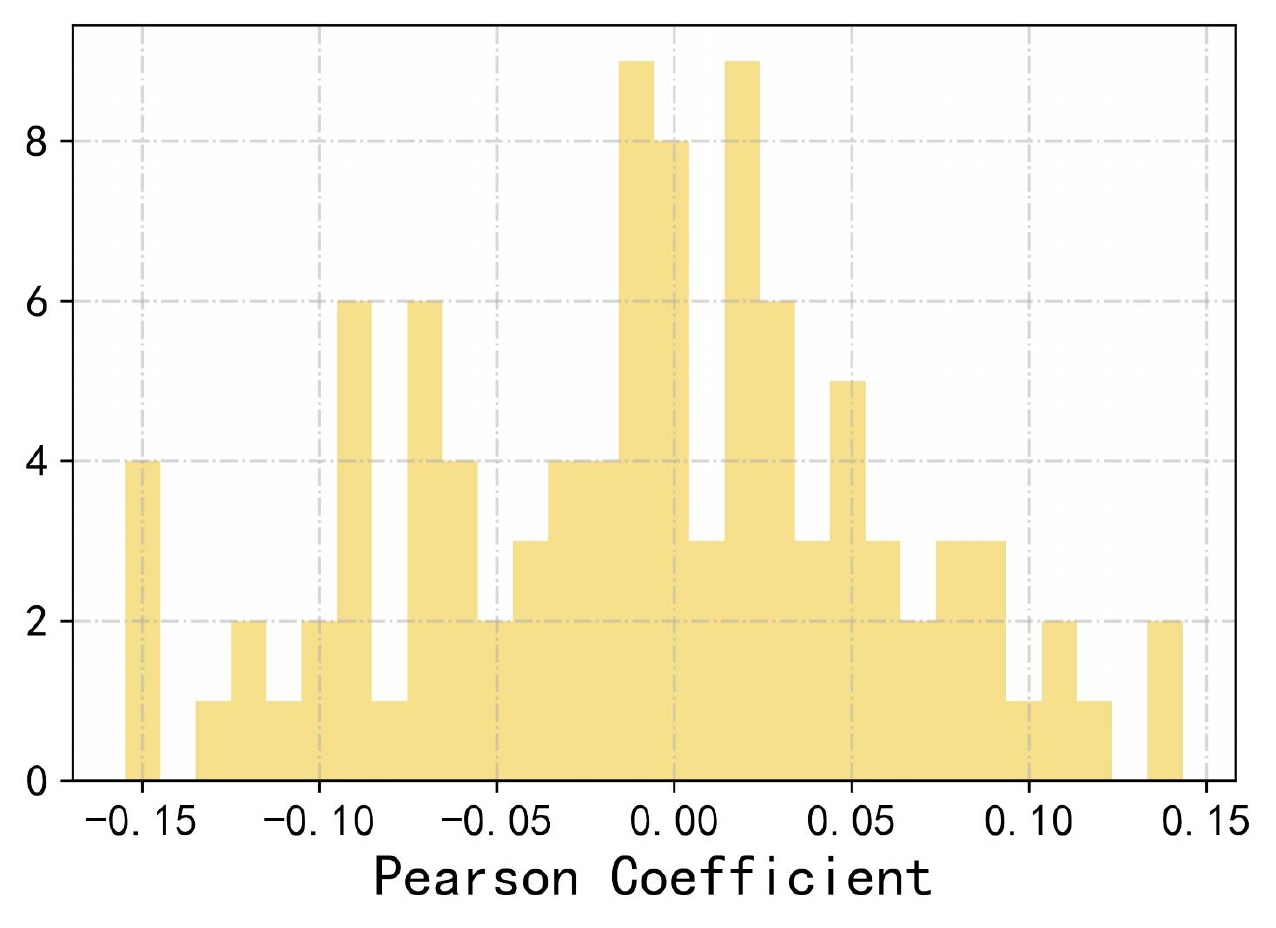}
	\end{minipage}} \\
	\caption{The frequency distribution of cancer driver genes with respect to correlation coefficients between the predicted differential expression levels and real ones, compared to random guess. The differential expression levels were predicted by slide-level pathological features on TCGA-BRCA cohort.}
	\label{fig:pred_vs_rand}
\end{figure}

For the regression task of differential expression levels, we used Pearson and Spearman correlation coefficient as evaluation metrics. On the TCGA-BRCA cohort, almost all genes yield significant prediction results. The average Pearson correlation coefficient of all test genes is 0.185 (p-value$<$0.01). Also, we observed inter-gene variation of the coefficients. Among the 200 tested genes, the Pearson coefficient of 84 genes is more than 0.20, and 39 genes is more than 0.40. The distribution of coefficients was shown in Figure~\ref{fig:pred_vs_rand} (a). To verify the reliability of the differential expression prediction, we compared our prediction results with random baseline. For this purpose, we generated random numbers within the 5\%-95\% range of the real differential expression (logarithm transformed fold changes) across all samples for each gene, and then calculated the correlation coefficient between the random baseline and the real differential expression levels. As shown in Figure~\ref{fig:pred_vs_rand} (a), we noticed the correlation coefficients followed nearly Gaussian distribution ($\mu$=0 and $\sigma$=0.15), which is far from the results of our prediction results. The statistical test verified the difference between our prediction and random generation is strongly significant ($p$-value=2.39e-17, Wilcoxon rank sum test).

We went further to run hypotheses test for each gene to check whether the predicted differential expression levels were significantly different from random baseline. The bilateral Wilcoxon test with Benjamin-Hochberg multiple testing correction was conducted for each gene. Among the 200 cancer driver genes, 88\% ($n$=176) genes were verified ($p$-value$<$0.05).

\begin{figure}[htb]
	\centering
	\includegraphics[width=0.8\columnwidth]{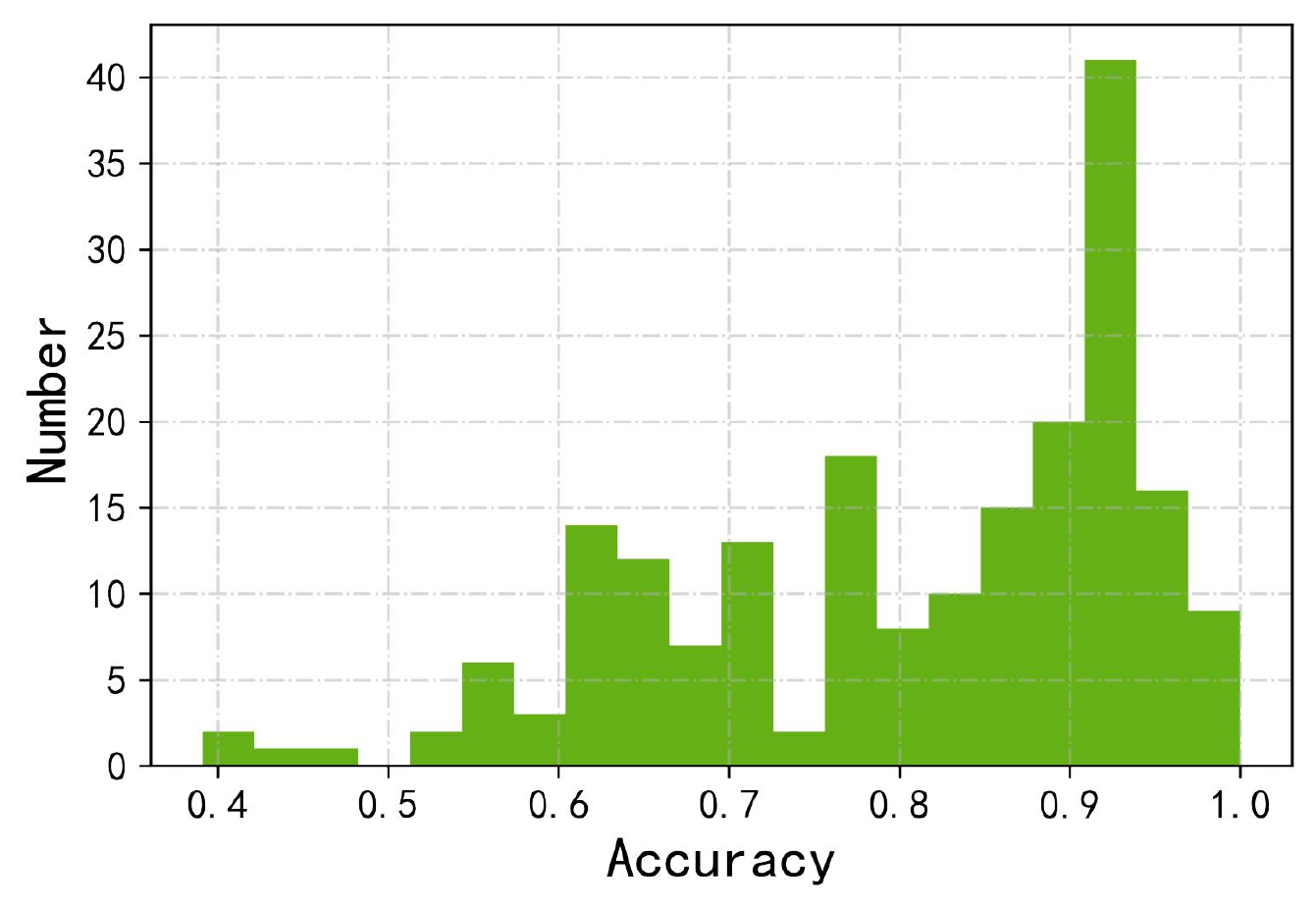}
	\caption{The frequency histogram of cancer driver genes with respect to their prediction accuracy of differential expression. The differential expression was cast to a binary classification task, and the prediction was based on the slide-level pathological features on TCGA-BRCA cohort.}
	\label{fig:DE_class_dist}
\end{figure}

Moreover, we cast the differential expression prediction to binary classification problem, by discretizing fold change levels. If the $|\log_2(fc)|$ was more than 1.5, the gene was regarded as significantly up-regulated or down-regulated and its label was set to 1, and 0 otherwise. Upon the same slide-level features and linear prediction model, we just replaced the output layer by two nodes and used softmax activation function for classification task. For most of 200 cancer driver genes, their differential expression can be significantly predicted, as shown in Figure~\ref{fig:DE_class_dist}. The predictive accuracy 0.9 account for more than 33 percent genes($n$=66).  There are 60 percent genes($n$=119), their prediction accuracy was more than 0.8. This result also supported that our method achieved notably performance in predicting differential gene expression using computational pathological features.

\subsection{Contrastive learning and attentive pooling improved performance}
\begin{figure}[htb]
	\centering
	\includegraphics[width=\columnwidth]{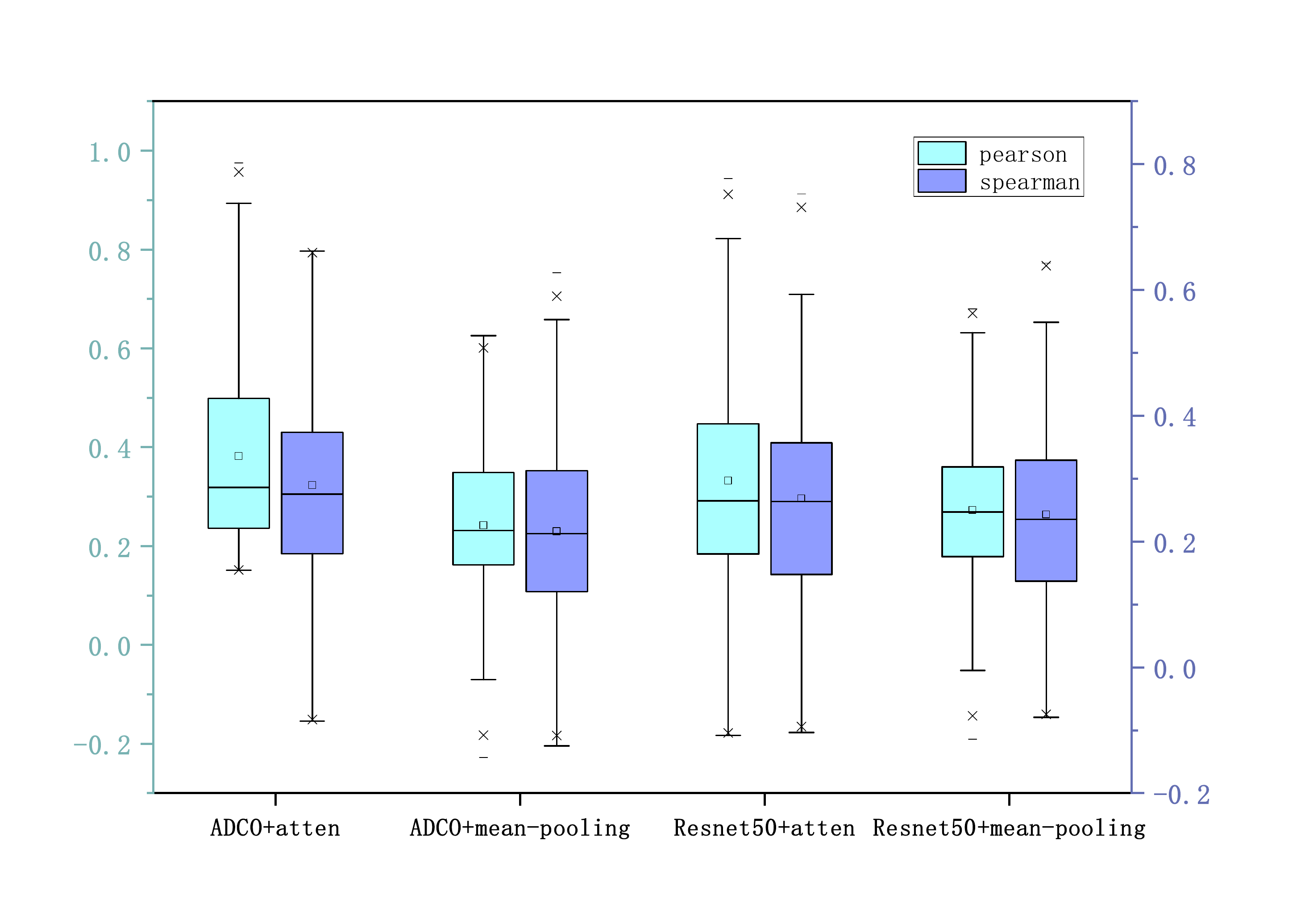}
	\caption{Comparison of the models with pretraining versus without pretraining, as well as attention pooling versus mean-pooling. The performance was evaluated on the differential expression prediction task. }
	\label{fig:ablation}
\end{figure}

To verify the contrastive learning-based feature extraction improve the performance of downstream task, we compared our method to supervised model without pretraining. For this purpose, we used the ResNet50 network pretrained on ImageNet as backbone to extract feature from tiles, and attention pooling was applied for feature aggregation. For the differential expression prediction task, we reported Pearson and Spearmen coefficients in Figure \ref{fig:ablation}. When attention pooling was applied, the AdCo-based contrastive learning  performed better than ResNet50-based feature extraction (AdCo+attention vs ResNet50+ attention).

Meanwhile, we investigated the impact of gated-attention mechanism. First, in the tumor diagnosis classification task, we have validated attention-pooling outperformed mean-pooling and max-pooling. Here, we verified its improvement of performance on regression task. Without loss of generality, we took into account contrastive learning-derived and supervised learning-derived tile-level feature. As shown in Figure~\ref{fig:ablation}, attention-pooling achieved better prediction accuracy than mean-pooling, for both AdCo-based and ResNet50-based features.

\subsection{Fold-change level positively correlated to prediction accuracy}
\begin{figure}[htb]
	\centering
	\includegraphics[width=0.8\columnwidth]{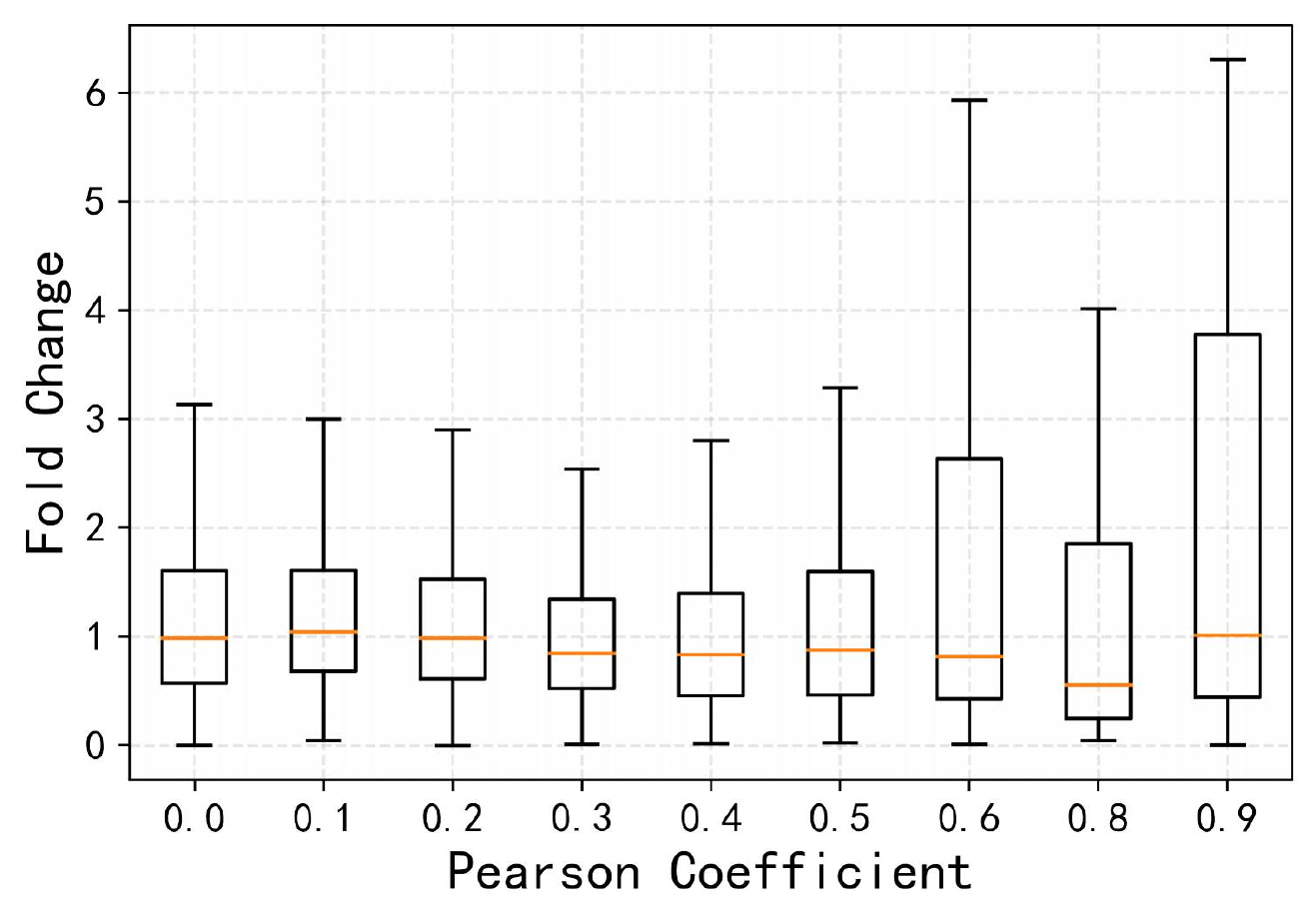}
	\caption{Boxplot of fold changes with respect to predictive performance of differentail expression. The cancer driver genes were binned according to the correlation coefficients of predicted and real differential expression levels.}
	\label{fig:fc_corr_boxplot}
\end{figure}

With the assumption that molecular pattern underlies the histological morphology, we speculated that the change of expression level can be reflected in the pathological feature. More intuitively, greater change of gene expression level leads to more significant pathological feature change that can be captured by our model. To verify this viewpoint, we checked the fold-change levels is correlated to the prediction accuracy of differential expression. Therefore, we drew boxplot of fold-change levels with respect to Pearson coefficients, as shown in Figure~\ref{fig:fc_corr_boxplot}. Overall, the prediction accuracy (Pearson coefficient) is positively correlated to the fold-change level. In particular, for the driver genes with prediction accuracy greater than 0.2, the mean fold changes was 2.602, while the genes with prediction accuracy less than 0.2, the mean fold change was only 2.198. For those genes with  prediction accuracy greater than 0.5, the mean fold-change level reached 3.241. Therefore, we drew the conclusion that our model actually captured the underlying molecular features dominated by gene expression patterns.

\subsection{Frequently mutated and immune-related genes were well predicted}
For further verification of our model, we checked whether the differential expression of frequently mutated driver genes are well predicted or not. Besides, as tumor immune microenmironment reflected the interaction between immune system and tumor evolution, we also considered the immune-related genes. From the list of ten frequently mutated genes in breast cancer\citep{nik2016landscape}, we got four genes, PIK3CA, MYC, PTEN and GATA3, which overlapped with compendium of cancer driver genes. We also selected four genes, CD3D, CD3E, CD3G, CD247, that encode the subunits of T lymphocyte glycoprotein CD3 receptor\citep{loh1987identification}. For B cell population, its marker gene CD19 was included.

\begin{figure}[htb]
	\centering
	\includegraphics[width=\columnwidth]{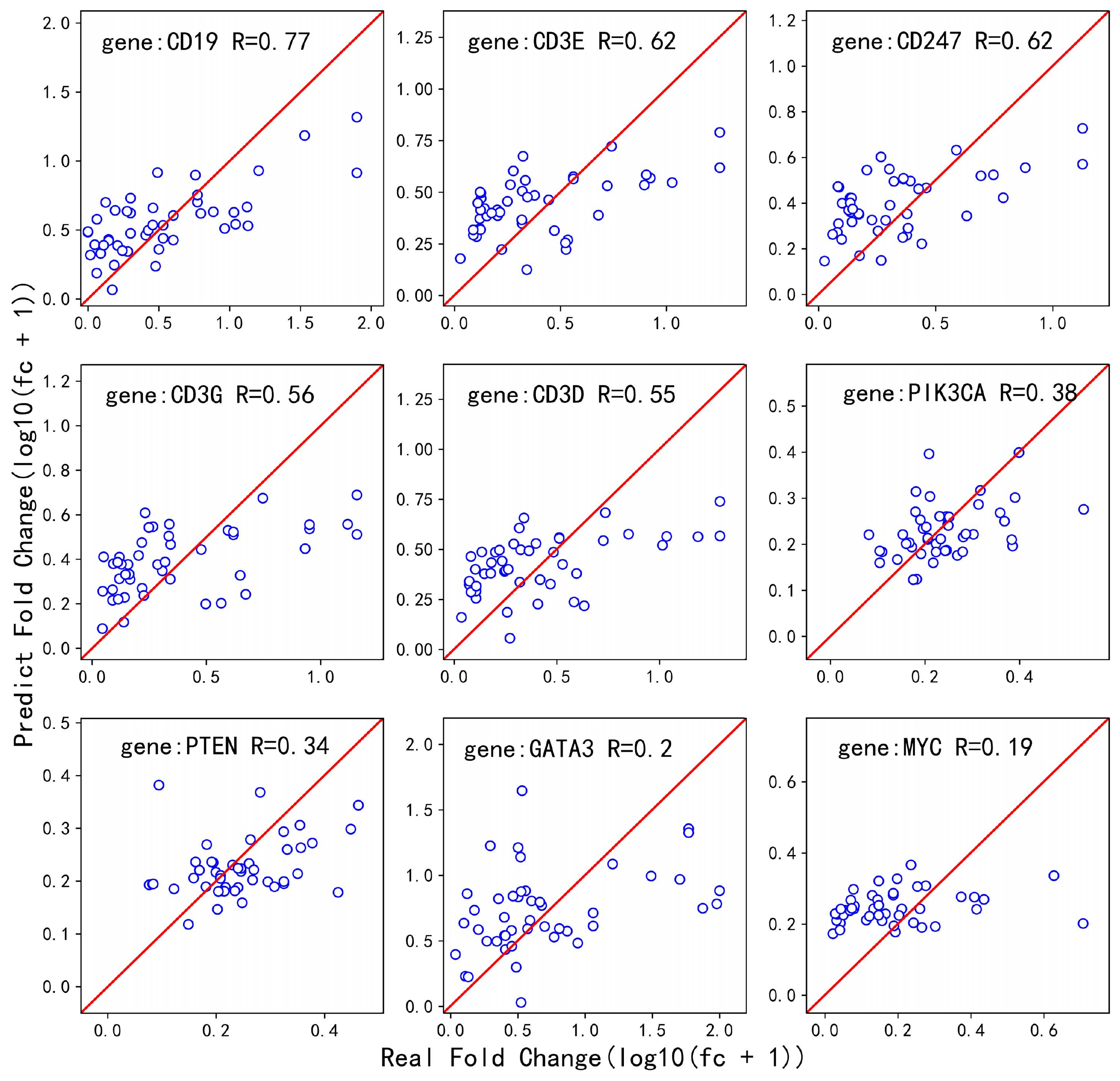}
	\caption{Scatter plots of predicted and real differential expression levels for 4 frequently mutated genes in breast cancer and 5 immune-related genes. For each gene, the Pearson coefficient was also shown.}
	\label{fig:top9_scatter}
\end{figure}

The results are shown in the figure~\ref{fig:top9_scatter}. It can be seen that the differential expression patterns of these frequently mutated and immune-related genes can be significantly predicted. In particular, the genes related to immune cells achieved the average Pearson correlation coefficient 0.624. For CD19 gene, the predicted fold changes showed strongly positive correlation to RNA-seq results.

For TCGA-BRCA cohort, we inspected the biological functions of top 5 best predicted genes among all 200 driver genes, including Elf3, SRGAP3, FGFR2, BRCA2 and FANCD2. Among them, BRCA2 ($p$-value=2.3e-7) is the notorious susceptibility gene to breast cancer and ovarian cancer\citep{wooster1995identification}, and involved in cell cycle control, gene transcription regulation, DNA damage repair, apoptosis and other important processes\citep{wang2012common}. Also, FGFR2 (p-value=4.8$e$-4) plays an important role in regulating cell proliferation, survival, migration and differentiation, and can induce mitosis and promote the occurrence of cancer\citep{wesche2011fibroblast}. Elf3 ($p$-value=1.2e-6) plays an important role in development, differentiation and transformation\citep{schedin2004esx}. SRGAP3 (p-value=7.9$e$-6) regulate cytoskeleton and participate in cell migration\citep{yang2006megap}. FANCD2($p$-value=5.7e-6)  protein is necessary to ensure efficient replication of common fragile sites\citep{madireddy2016fancd2}.

\begin{figure}[htb]
	\centering
	\includegraphics[width=0.85\columnwidth]{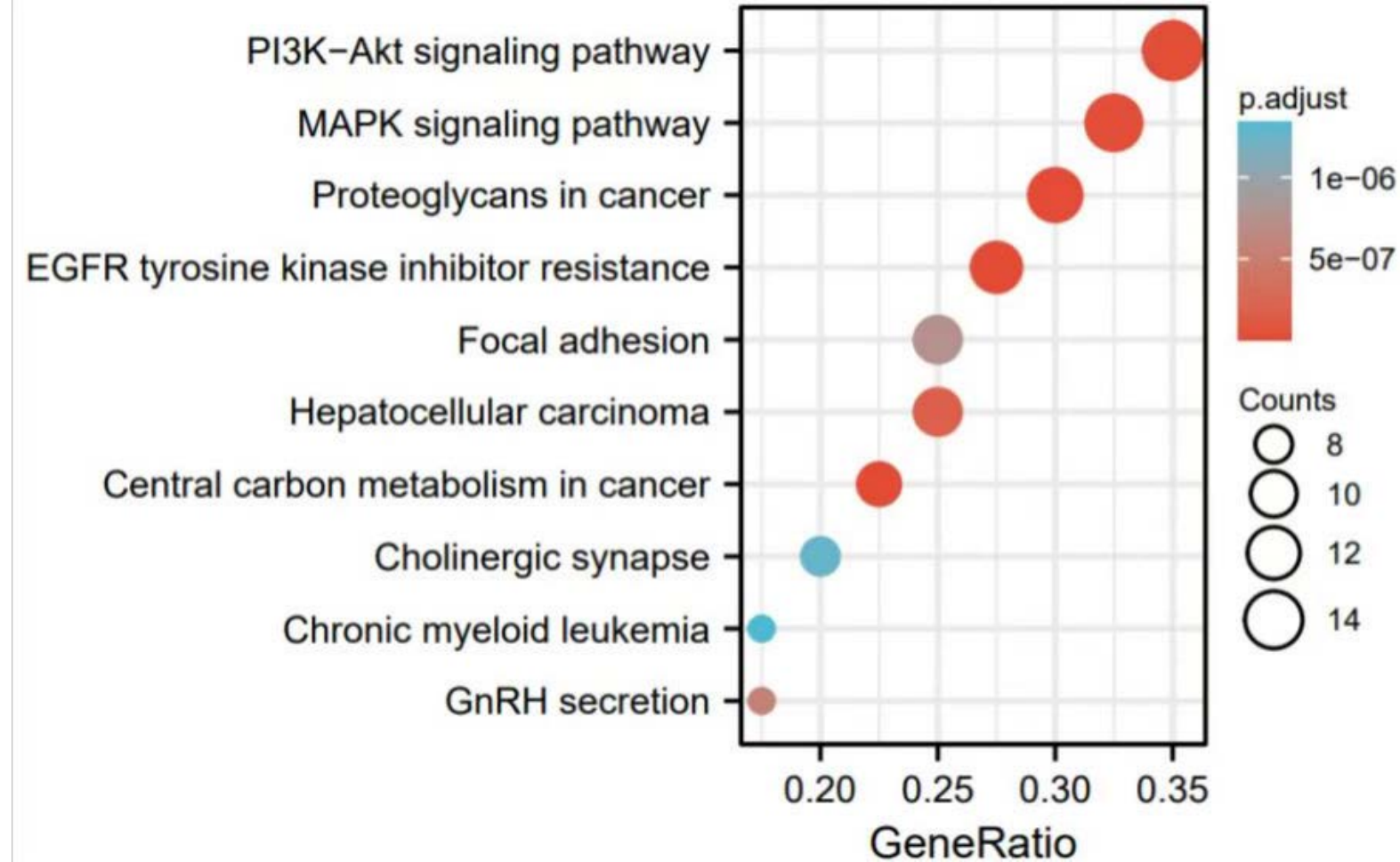}
	\caption{Enrichment analysis on KEGG signaling pathway based on top 50 genes of which differential expression were predicted.}
	\label{fig:kegg}
\end{figure}

Finally, we selected 50 most predicted genes for enrichment analysis. The results are shown in Figure \ref{fig:kegg}. Among the top 10 enriched signaling pathways, we found the PI3K pathway, MAPK pathway and EGFR signaling that have been reported abnormal activation in breast cancer.

\subsection{Interpretability and spatial localization}
\begin{figure*}[htb]
	\centering
	\includegraphics[scale=0.85]{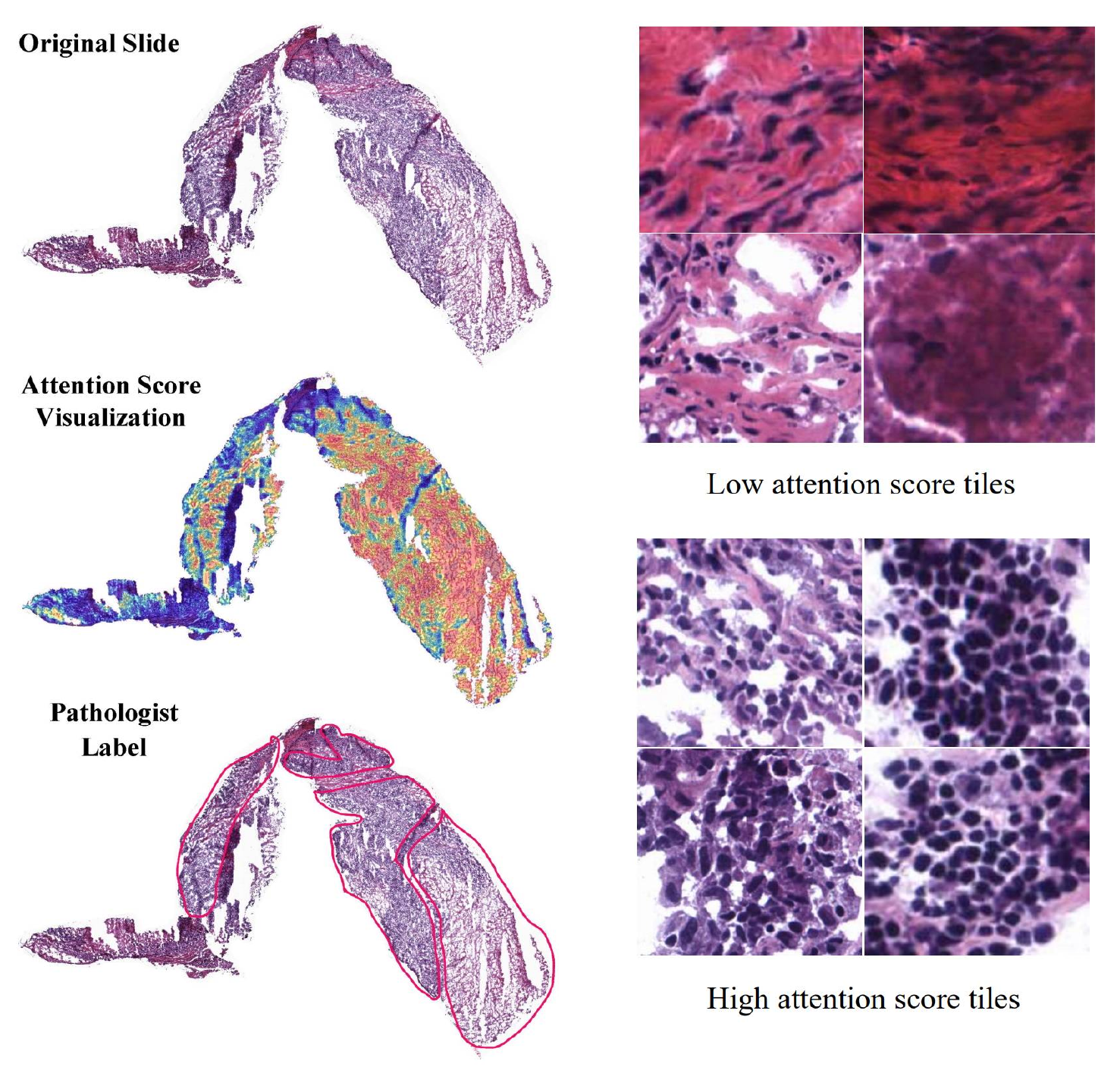}
	\caption{Spatial localization of tumor area of implied by attention mechanism compared to pathologist annotations. The heatmap (left middle) was generated by spatial deconvolution of tiles to original slide (left top), and each tile was colored according to its normalized attention score. The area circled by red lines were tumor\&necrosis area (left bottom) annotated by an experienced pathologist. The right column showed some representative tiles with highest and lowest attention scores.}
	\label{fig:pred_vs_annot}
\end{figure*}

Human-readable interpretability of the extracted feature from pathology images can validate that the predictive power of our model kept in line with well-known morphology annotated by experienced pathologist. Our slide-level prediction of tumor diagnosis was made by identifying and aggregating tile-level features that are of high diagnostic importance (high attention score). The derived attention scores are highly indicative of localization of tumor area. To visualize and interpret the relative importance of each region of the WSIs, we generated the attention heatmap by normalization of attention scores and spatial deconvolution of tiles to original slide. Fine-grained attention heatmap were evaluated by aligning to annotated tumor and necrosis, normal areas by pathologist. As shown in Figure~\ref{fig:pred_vs_annot}, the attention heatmap reflected the spatial localization of highly diagnostic tissues, which were coincident with the boundary of tumor\&necrosis and normal tissues delineated by the pathologist. Visiting tiles of both high and low attention scores within a slide would convey more human-readable pathological feature. So, we visualized a few tiles selected according attention scores, and found that the tiles with low attention scores are mostly normal tissue, while those with high attention scores were dominately covered by tumor cells.

Beyond the spatial localization of tumor tissue, differential gene expression also allowed us to visualize the spatial distribution of immune infiltrating cells. For this purpose, we chose the immune cell marker genes for further analysis. As T lymphocytes and B lymphocytes infiltrating to tumor tissue play main anti-cancer effect, we selected several genes specific to T and B lymphocytes. For T lymphocyte, we chose CD3D and CD247 genes that encode T lymphocyte receptor glycoprotein CD3. For B lymphocyte, the CD19 gene was chosen. Besides, we used another tumor suppressor gene PTEN for comparison. For each gene, we generated the slide-level heatmap using the attention scores of each tiles derived from the differential expression prediction models specific to this gene. Meanwhile, we asked the pathologist to manually label the tumor area, which was used as reference boundary of the spatial distribution of immune infiltrating cells.

\begin{figure*}[htbp]  
	\captionsetup{labelformat=simple, position=top}
	\centering
	\subfloat[Annotation]
	{
		\begin{minipage}[b]{.4\columnwidth}
			\centering
			\includegraphics[width=3.5cm]{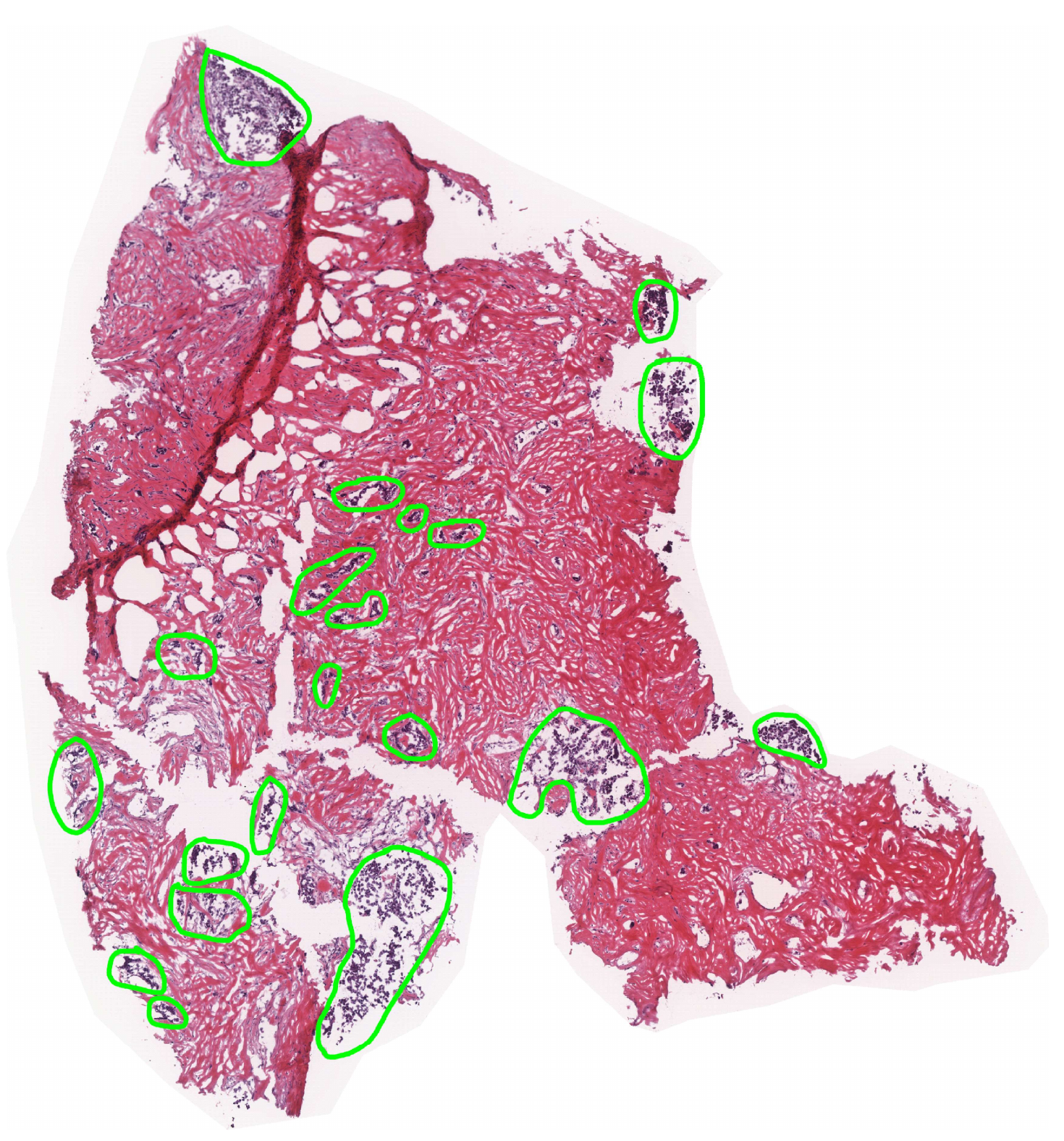}
	\end{minipage}}
	\subfloat[CD3D ]
	{
		\begin{minipage}[b]{.4\columnwidth}
			\centering
			\includegraphics[width=3.5cm]{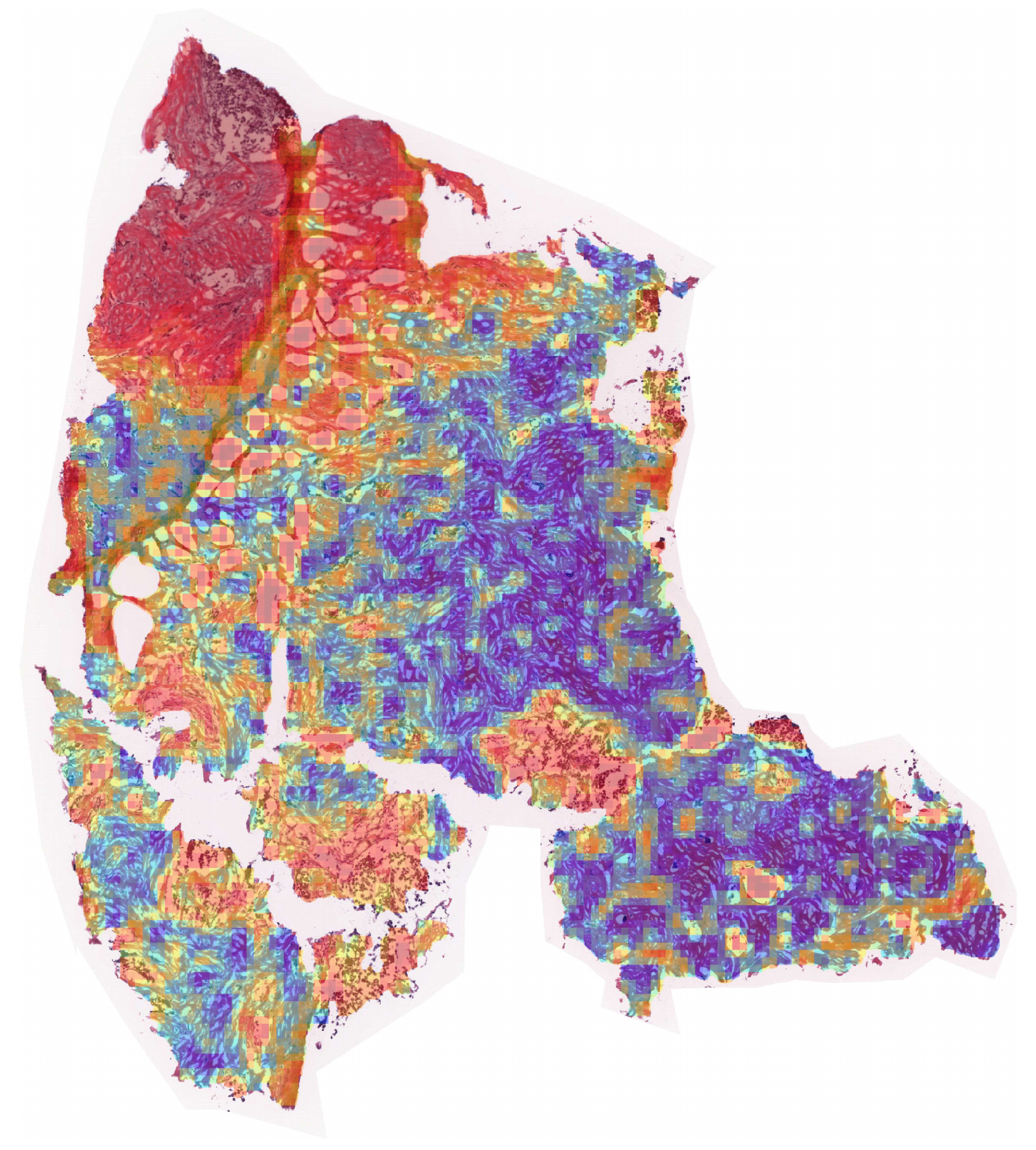}
	\end{minipage}}
	\subfloat[CD247 ]
	{
		\begin{minipage}[b]{.4\columnwidth}
			\centering
			\includegraphics[width=3.5cm]{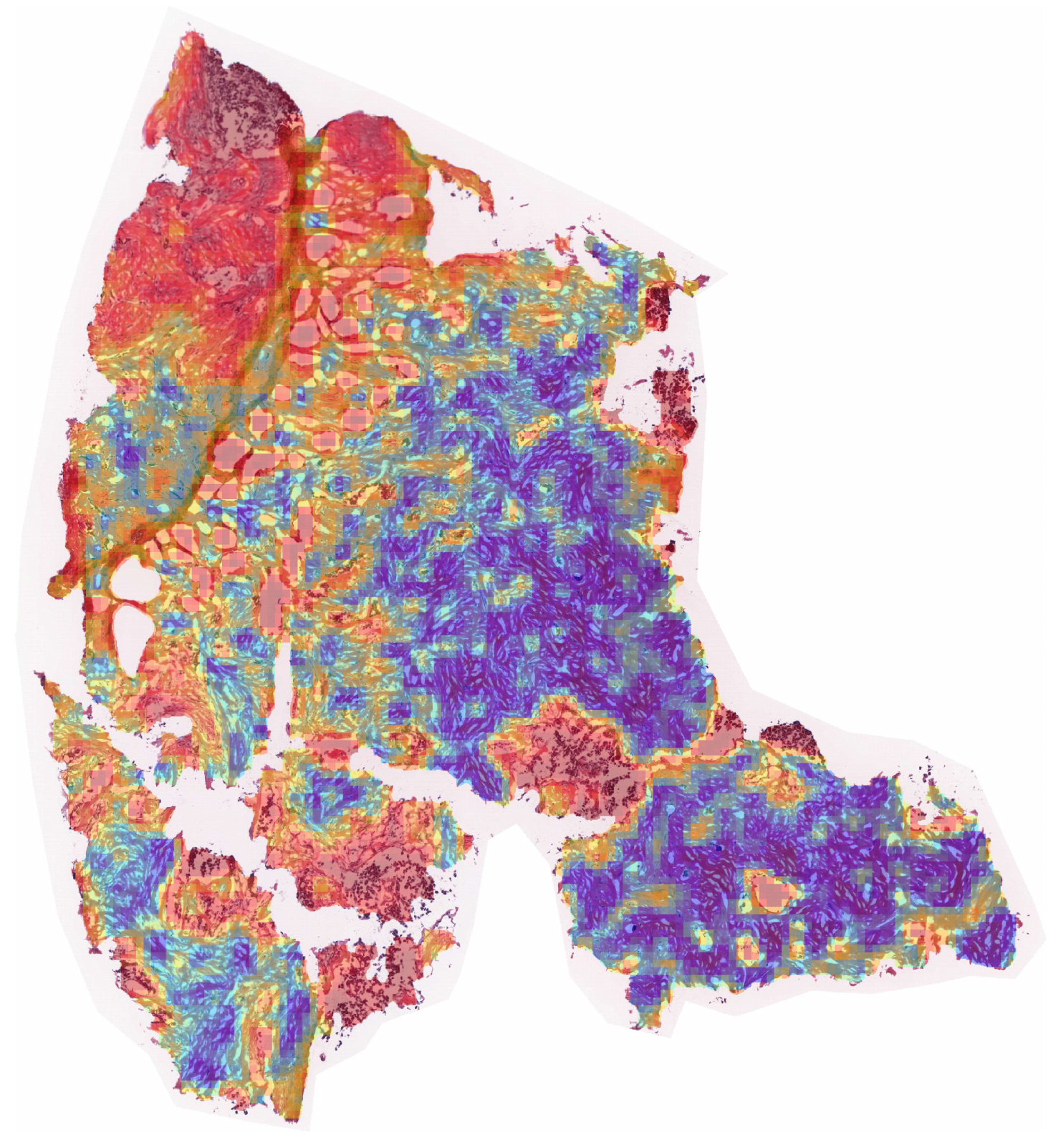}
	\end{minipage}}
	\subfloat[CD19 ]
	{
		\begin{minipage}[b]{.4\columnwidth}
			\centering
			\includegraphics[width=3.5cm]{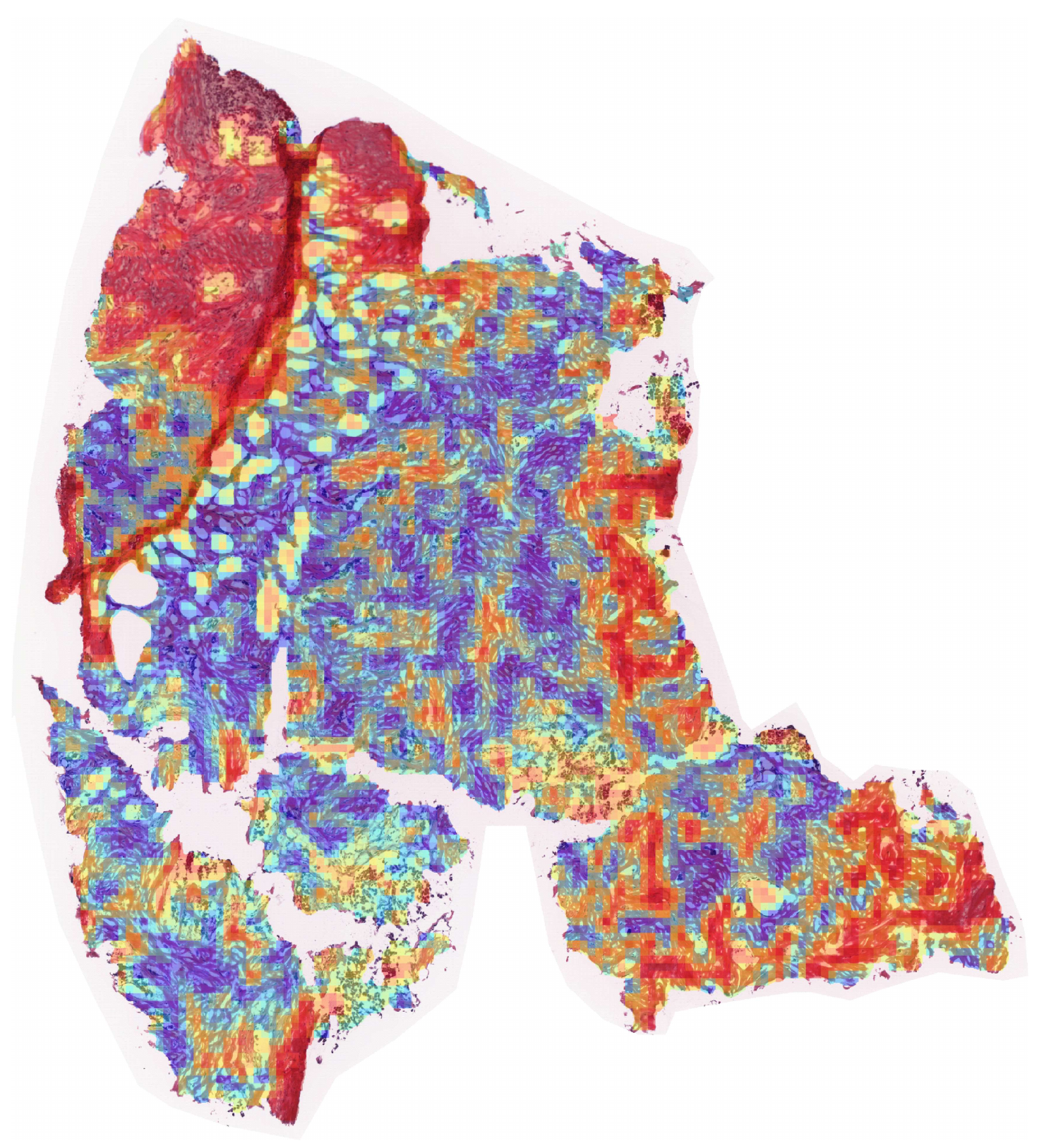}
	\end{minipage}}
	\subfloat[PTEN ]
	{
		\begin{minipage}[b]{.4\columnwidth}
			\centering
			\includegraphics[width=3.5cm]{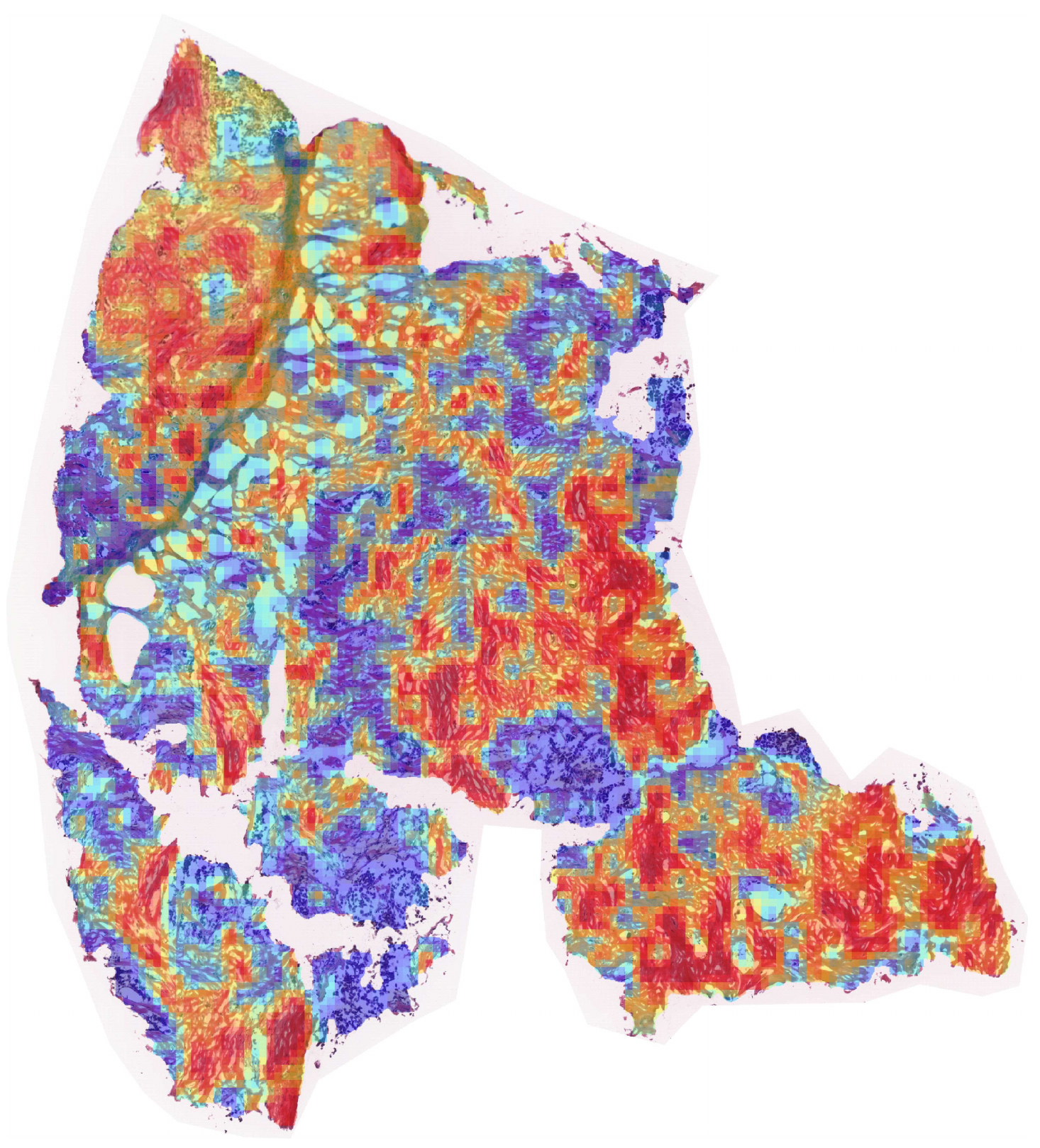}
	\end{minipage}}
	\caption{Spatial localization of tumor area and cells labeled by marker genes. The tumor area was labeled by an experienced pathologist. The localization of T lymphocytes and B lymphocytes was predicted by their marker genes.}
	\label{fig:genes_heatmap}
\end{figure*}

As shown in Figure \ref{fig:genes_heatmap}, the heatmap of CD3D and CD247 marker genes indicated that T lymphocyte mostly located at the tumor area. For the CD19 gene reflecting B lymphocyte localization, we observed similar spatial distribution to T lymphocyte. The observation was consistent to the fact that immune cells infiltrated to solid tumor to kill cancer cells by recognition of neoantigen. In contrast, the PTEN expression showed quite different spatial distribution. Interestingly, we found that PTEN gene showed high expression levels in normal tissue relative to tumor tissue.

\section{Discussion and Conclusion}
Genetic testing has been an important clinical examination for tumor subtyping and targeted drug delivery. However, due to technology and cost reasons, genetic testing has not become clinically spread, especially in developing countries. The rapid advancement of digital pathology motivated the application of computational pathology to infer genetic mutations, microsatellite instability and tumor microenvironment. Following the rationale that the change of cell and tissue phenotype is driven by the variation of gene expression pattern, we have explored the computational pathology-based prediction of differential expression of cancer driver genes. In general, the hematoxylin and eosin-stained digital slides contained both tumor and normal tissues, this allow us to infer differentially expressed genes from pathological feature from single whole slide image. In fact, we have shown that the fold-change levels positively correlated to prediction accuracy. This indicated that dramatic variation of underlying molecule expression pattern would be more reflected in phenotypic features, which in turn allow us to infer differential expression.

The inference from pathological feature to molecule patterns mainly depends on the powerful feature extraction capacity of deep learning. Although weakly-supervised deep learning has been applied in a few studies for computational pathology, self-supervised contrastive learning showed superior performance in representation learning, but has not been exploited in mining digital pathology. Our study verified the contrastive learning-based pretraining significantly improved the performance in downstream classification (tumor diagnosis) and regression (differential expression) tasks. We speculate that contrastive learning can capture fine-grained information in learning to acquire similar features for tile-level positive pair but dissimilar from a set of negative tiles. In contrast, weakly-supervised learning may only extract information for slide-level instance classification.

The bulk RNA-seq dataset is the standard for differential expression analysis, which has been widely used to identify up-regulated and down-regulated genes in tumor cells compared to normal cells. However, bulk sequencing lost the spatial localization of tumor and normal tissues. The spatial deconvolution of marker gene expression to histopathology images greatly help to visualize the spatial distribution of tumor and normal cells, especially the immune infiltrating cells. Looking forward, computational pathology would promote the spatial visualization of tumor immune microenvironment.

In summary, we developed a self-supervised contrastive learning framework, HistCode, to infer differential gene expression from pathology images. Our extensive experiments showed that contrastive learning-based pretraining effectively improved the downstream task, including tumor diagnosis and differential expression prediction. We have also shown informative spatial visualization of tumor and specific gene expression by leveraging the tile-level attention scores learned by our model. We believe that our study would yield inspiring insight into computational pathology.

\section{Author contributions statement}
H.H. and H.L. conceived the main idea and the framework of the manuscript. H.L. collected the datasets. H.H. drafted the manuscript. H.H. and G.Z. performed the experiments. X.L. and L.D. helped to improve the idea. C.W. reviewed drafts of the paper. D.Z. annotated the pathology images. H.L. revised the manuscript. H.L. supervised the study and provided funding. All authors read and commented on the manuscript.

\section{Data availability}
Source code and all datasets used in this study are available at \\ \href{https://github.com/hoarjour/HistCode/}{https://github.com/hoarjour/HistCode/}

\section{Acknowledgments}
We would like to thank The Cancer Genome Atlas (TCGA) for providing free WSI datasets and RNA-seq data. This work was supported by the National Natural Science Foundation of China under grants No.~62072058 and No.~61972422.

\bibliographystyle{unsrt}
\bibliography{reference}



\begin{biography}{}{\author{Haojie Huang} is an undergraduate student at School of Computer Science and Engineering, Central South University, Changsha, China.}
\end{biography}
\begin{biography}{}{\author{Gongming Zhou} is an undergraduate student at School of Computer Science and Engineering, Central South University, Changsha, China.}
\end{biography}
\begin{biography}{}{\author{Xuejun Liu} is a professor at School of Computer Science and Technology, Nanjing Tech University, Nanjing, China. His research interests include data mining and deep learning.}
\end{biography}
\begin{biography}{}{\author{Lei Deng} is a professor at School of Computer Science and Engineering, Central South University, Changsha, China. His research interests include data mining, bioinformatics and systems biology.}
\end{biography}

\begin{biography}{}{\author{Chen Wu} is experienced oncologist at The third affiliated hospital of Soochow University, 213100, Changzhou, China. His research interest include cancer immune-therapy.}
\end{biography}

\begin{biography}{}{\author{Dachuan Zhang} is an experienced pathologist at The third affiliated hospital of Soochow University, 213100, Changzhou, China. His research interest include Tumor immune microenvironment.}
\end{biography}
\begin{biography}{}{\author{Hui Liu} is a professor at School of Computer Science and Technology, Nanjing Tech University, Nanjing, China. His research interests include Bioinformatics and Deep Learning.}
\end{biography}

\end{document}